\documentclass[10pt,journal]{IEEEtran}

\usepackage{cite}
\usepackage[pdftex]{graphicx}

\usepackage{caption}
\usepackage{subcaption}

\ifCLASSINFOpdf
   \usepackage[pdftex]{graphicx}
\else
   \usepackage[dvips]{graphicx}
\fi
\usepackage{amsmath}
\usepackage{amssymb}
\usepackage[noend]{algpseudocode}
\usepackage{algorithm}

\usepackage{array}
\begin{document}
%
\title{Algorithm for Decentralized Cooperative Positioning of Multiple Autonomous Agents}
%
%
%

\author{Milutin~Pajovic,~\IEEEmembership{Member,~IEEE,}
        Vikrant~Shah,~\IEEEmembership{Member,~IEEE,}
        and~Philip~V.~Orlik,~\IEEEmembership{Senior Member,~IEEE}
\thanks{M. Pajovic and P. V. Orlik are with the Mitsubishi Electric Research Laboratories (MERL), Cambridge, MA, 02138, USA. Email: pajovic@merl.com}
\thanks{V. Shah is with the Computer Engineering Department, Northeastern University,
        Boston, MA, 02115, USA. This work has been performed when he was an intern at MERL.}}
\maketitle

\begin{abstract}
One of the most essential prerequisites behind a successful task execution of a team of agents is to accurately estimate and track their poses. We consider a cooperative multi-agent positioning problem where each agent performs single-agent positioning until it encounters some other agent. Upon the encounter, the two agents measure their relative pose, and exchange particle clouds representing their poses. We propose a cooperative positioning algorithm which fuses the received information with the locally available measurements and infers an agent's pose within Bayesian framework. The algorithm is scalable to multiple agents, has relatively low computational complexity, admits decentralized implementation across agents, and imposes relatively mild requirements on communication coverage and bandwidth. The experiments indicate that the proposed algorithm considerably improves single-agent positioning accuracy, reduces the convergence time of a particle cloud and, unlike its single-agent positioning counterpart, exhibits immunity to an impeding feature-scarce and symmetric environment layout.
\end{abstract}


\section{Introduction}


Positioning is at the core of numerous applications supported by autonomous agents. It is often a challenging task, especially in environments without Global Positioning System (GPS) coverage and/or in setups where agents use sensors of limited capabilities. In cooperative positioning (CP), multiple agents cooperate with the goal to estimate their poses (i.e., locations and orientations) more accurately and thus meet strict positioning accuracy requirements. As such, the CP may be well suited for positioning of vehicles in Intelligent Transportation Systems (ITS)~\cite{CITS_1,CITS_2}, especially in environments such as urban canyons~\cite{urban_canyon} where vehicles' cooperation is expected to overcome issues arising from the lack of full GPS coverage. Consequently, different cooperative schemes have been considered to enhance GPS-based position estimates~\cite{lit_CP_GPS_1,lit_CP_GPS_2,lit_CP_GPS_3,lit_CP_GPS_4,lit_CP_GPS_5}. 
The initial GPS locations are improved with vehicle cooperation, realized through detection and localization of common features using data association in~\cite{BayesianCP_PDA_separate}~\cite{BayesianCP_PDA_joint}.

In another application, the CP is a critical part of cooperative multi-agent Simultaneous Localization and Mapping (SLAM), where agents' cooperation provides benefits such as faster convergence and improved robustness, in addition to obtaining more accurate estimates of the map and agents' poses~\cite{Saeedi_review_16}. In the robotics community,~\cite{red_odom_err_1,red_odom_err_2,red_odom_err_3} exploit robot cooperation to reduce odometry errors,~\cite{opt_based_1,opt_based_2} frame the CP as a nonlinear least squares optimization problem, while~\cite{Kalman_1,Kalman_2,Kalman_3} are Kalman filter-based solutions with different levels of decentralization and communication bandwidth requirements.
Particle filtering (PF) is at the core of state of the art algorithms for single-robot positioning~\cite{AMCL_alg} and SLAM~\cite{SLAM}, and has also been used for tracking multiple robots that do not make relative pose measurements~\cite{CL_tracking}, as well as in multi-robot SLAM~\cite{Howard_MultiRobotSLAM_06,Carlone_SLAM_10} where robots exchange all measurements at each encounter.

In this paper, we propose an algorithm for cooperative multi-agent pose estimation and experimentally validate it. The agents in our setup perform pose estimation on their own using PF. At an encounter, the agents detect the presence of each other, measure relative pose between them, exchange pose particles, and update the pose particles using the proposed algorithm. 
The algorithm is built upon the Bayesian inference framework, is implemented as a fully decentralized PF, and has computational complexity of the order of the conventional single-agent positioning algorithm~\cite{AMCL_alg}. The experimental study shows that the particles in the proposed algorithm quickly converge, achieve a considerably better accuracy compared to the single-agent positioning without cooperation and, most importantly, overcome the impediments arising from a symmetric and feature scarce experimental area. Notably, the proposed method does not rely on GPS signals and is thus suited for GPS-denied environments. 

Among the CP algorithms,~\cite{Fox_MultiRobotLocalization_00} and~\cite{Prorok_IROS_11} are the most relevant to our work as they build upon the PF framework and consider the same setup. While the information fusion in~\cite{Fox_MultiRobotLocalization_00} is aided with the use of density trees to infer an agent's pose from its particles, the particle updating scheme in~\cite{Prorok_IROS_11} comprises of evaluating relative pose distribution for all pairs of particles from two agents involved in the encounter, which results in $O(K^2)$ computational complexity, where $K$ is the number of particles. This complexity is reduced in~\cite{Prorok_ICRA_12} to $O(KS)$ by clustering $K$ particles into $S$ clusters. In comparison to these works, our CP algorithm is derived by properly abiding with the Bayesian inference principles and the resulting particle updating scheme is optimal in the Bayesian sense. As such, it does not employ density trees, directly fuses information obtained at the agents' encounter and has linear complexity in the number of particles, i.e, $O(K)$.

%

\section{Cooperative Multi-agent Positioning} \label{Algorithm}

We consider a scenario where multiple agents move in an indoor area whose map is known and made available to them. Each agent infers its own pose until it encounters other agent. At the encounter, the agents exchange information over a wireless link, and each agent fuses its own estimates with the information received from the other agent. 
In the following, we first outline a PF-based single-agent positioning algorithm and then derive the proposed CP algorithm.

\subsection{Single-Agent Pose Estimation} \label{subsec:singe_agent_pose}

A pose estimation of an agent in a 2D environment is concerned with estimating agent's 2D coordinates $(x,y)$ and orientation $\theta$ with respect to the coordinate system associated with the environment. The 2D coordinates and orientation at time $t$ are collected into a pose vector ${\bf x}_t=\left[ \begin{array}{ccc} x_t & y_t & \theta_t \end{array} \right]^T$, where $^T$ denotes vector/matrix transpose operator. The environment is represented with an occupancy grid map ${\bf m} \in \{0,1\}^N$, obtained by dividing the area into $N$ bins, such that $m_i=1$ in the case the $i$th bin is occupied, or $m_i=0$, otherwise. As the agent moves through the environment, it collects ranging and odometry measurements. The ranging measurement at time $t$ represents distances between the agent and obstacles seen within its field of view. They are collected into vector ${\bf z}_t$ and probabilistically modelled as $p({\bf z}_t | {\bf x}_t , {\bf m})$~\cite[Ch. 6]{Thrun_book_05}. The odometry measurements corresponding to time interval $(t-1,t]$ are collected into vector ${\bf u}_t$ and modelled as $p({\bf x}_t | {\bf x}_{t-1},{\bf u}_t)$~\cite[Ch. 5]{Thrun_book_05}.

The pose estimation within Bayesian framework comprises of inferring probability distribution of an agent's pose ${\bf x}_t$, given the map and all ranging and odometry measurements up to time $t$,
\begin{equation}
p({\bf x}_t | {\bf u}_{1:t}, {\bf z}_{1:t}, {\bf m}) \triangleq p({\bf x}_t | {\bf u}_1, \ldots, {\bf u}_t, {\bf z}_1, \ldots, {\bf z}_{t}, {\bf m}) \label{single_robot_pose}
\end{equation}
Assuming the Markov property, the inference of~(\ref{single_robot_pose}) is performed sequentially using Bayes' filter where belief about agent's pose at time $t$ is updated from the belief of its pose at time $t-1$ and ranging and odometry measurements collected between $t-1$ and $t$~\cite[Ch. 8]{Thrun_book_05}. In most general case of non-Gaussian noise and/or non-linear measurement models, the Bayes' filter is implemented as a particle filter (PF)~\cite{Ristic_book_04}, which represents the distribution~(\ref{single_robot_pose}) with $K$ pose particles $\hat{\bf x}_{t,k}$, so that with a slight abuse of notation,
\begin{equation}
p({\bf x}_t | {\bf u}_{1:t}, {\bf z}_{1:t}, {\bf m}) \approx \frac{1}{K} \sum_{k=1}^K \delta({\bf x}-\hat{\bf x}_{t,k}),
\end{equation}
where $\delta(\cdot)$ is the Dirac's delta function. In the simplest implementation, the PF is initialized with $K$ particles, uniformly sampled from the area where the agent is present. Given the set of particles $\{\hat{\bf x}_{t-1,k}\}_{k=1}^K$ representing the agent's pose at time $t-1$ and odometry measurement ${\bf u}_t$, the agent motion model $p({\bf x}_t | {\bf x}_{t-1}, {\bf u}_t)$ is used to sample (tentative) particles $\{{\bf x}^{\prime}_k\}_{k=1}^K$. Each tentative particle ${\bf x}^{\prime}_k$ is associated with a weight $w_k$ computed from the ranging model and measurement ${\bf z}_t$,
\begin{equation}
w_k \propto p({\bf z}_t | {\bf x}^{\prime}_k , {\bf m}) \label{single_robot_weights}
\end{equation}
and normalized so that $\sum_{k=1}^K w_k = 1$. Finally, the tentative particles are resampled according to $\{w_k\}_{k=1}^K$ to produce the particle set $\{\hat{\bf x}_{t,k}\}_{k=1}^K$ representing the agent's pose at time $t$.

\subsection{Cooperative Multi-Agent Pose Estimation}

Each agent moves through an area on its own until it encounters some other agent, at which point both of them cooperatively estimate/update their poses. In other words, the CP happens at agents' encounters. 
Irrespective of whether an encounter between two agents is prearranged or occurs as a result of their random wander, at least one agent involved in an encounter has to be able to detect the presence of the other one. Once the agents detect that they are in the vicinity of each other, one (or both) measure their relative pose ${\bf r}$, comprised of relative range $r$ and relative heading $\phi$ such that ${\bf r}=\left[\begin{array}{cc} r & \phi  \end{array} \right]^T$. There are various modalities as to how agents encounter each other and measure their relative pose; one experiments use one particular modality.

Assuming two agents $A$ and $B$ encounter each other at time $t$, our goal is to estimate their respective poses ${\bf x}^A_t$ and ${\bf x}^B_t$ based on their ranging and odometry measurements made prior to $t$, and their relative pose ${\bf r}_t$ measured at the moment of encounter. To aid pose estimation, we first consider joint probability distribution of both agents' traversed paths, conditioned on all measurements,
\begin{equation}
p \triangleq p({\bf x}^A_t, {\bf x}^B_t, {\bf x}^A_{1:t^-}, {\bf x}^B_{1:t^-} \; | \; \mathcal{D}^A_{1:t^-}, \mathcal{D}^B_{1:t^-}, {\bf r}_t ) \label{prob_dist_paths} 
\end{equation}
where $\mathcal{D}^A_{1:t^-} = \{ {\bf u}^A_{1:t^-}, {\bf z}^A_{1:t^-} \}$ and ${\bf x}^A_{1:t^-}$ respectively denote measurements (odometry and ranging) and traversed path corresponding to agent $A$, prior to time $t$. Analogous notation $\mathcal{D}^B_{1:t^-} = \{ {\bf u}^B_{1:t^-}, {\bf z}^B_{1:t^-} \}$ and ${\bf x}^B_{1:t^-}$ corresponds to agent $B$. We note that an agent does not change its pose from $t^-$ to $t$, nor takes odometry/ranging measurements. However its pose at $t^-$ is estimated from $\mathcal{D}^{A/B}_{1:t^-}$, while that at $t$ leverages additional information received from the encounter.

The joint path distribution $p$ in~(\ref{prob_dist_paths}) is using Bayes' rule expressed as
\begin{eqnarray}
\nonumber & p \propto p({\bf r}_r | {\bf x}^A_t, {\bf x}^B_t, {\bf x}^A_{1:t^-}, {\bf x}^B_{1:t^-}, \mathcal{D}^A_{1:t^-},\mathcal{D}^B_{1:t^-}) \times \\
& p({\bf x}^A_t, {\bf x}^B_t, {\bf x}^A_{1:t^-}, {\bf x}^B_{1:t^-} | \mathcal{D}^A_{1:t^-}, \mathcal{D}^B_{1:t^-}) \label{prob_dist_paths_1}
\end{eqnarray}
Since the relative pose ${\bf r}_t$ directly depends on ${\bf x}^A_t$ and ${\bf x}^B_t$, which already contain information about poses prior to $t$ and odometry/ranging measurements,
\begin{equation}
p({\bf r}_r | {\bf x}^A_t, {\bf x}^B_t, {\bf x}^A_{1:t^-}, {\bf x}^B_{1:t^-}, \mathcal{D}^A_{1:t^-},\mathcal{D}^B_{1:t^-}) = p({\bf r}_r | {\bf x}^A_t, {\bf x}^B_t) \label{prob_dist_paths_term_1}
\end{equation}
Conditioned on odometry/ranging measurements, agents' paths are independent of each other and hence the second term on the right side of~(\ref{prob_dist_paths_1}) factorizes as
\begin{eqnarray}
\nonumber & p({\bf x}^A_t, {\bf x}^B_t, {\bf x}^A_{1:t^-}, {\bf x}^B_{1:t^-} | \mathcal{D}^A_{1:t^-}, \mathcal{D}^B_{1:t^-}) = \\
& p({\bf x}^A_t, {\bf x}^A_{1:t^-} | \mathcal{D}^A_{1:t^-} ) \; p({\bf x}^B_t, {\bf x}^B_{1:t^-} | \mathcal{D}^B_{1:t^-} ) \label{prob_dist_paths_1_fact}
\end{eqnarray}
Recalling that an agent does not move nor takes ranging measurements between $t^-$ and $t$, the path distribution of agent $A$ is computed as
\begin{eqnarray}
\nonumber p({\bf x}^A_t, {\bf x}^A_{1:t^-} | \mathcal{D}^A_{1:t^-} ) &=& p({\bf x}^A_t | {\bf x}^A_{t^-}) \; p({\bf x}^A_{1:t^-} | \mathcal{D}^A_{1:t^-}) \\
&=& \delta({\bf x}^A_t-{\bf x}^A_{t^-}) \; p({\bf x}^A_{1:t^-} | \mathcal{D}^A_{1:t^-}) \label{path_dist_A}
\end{eqnarray}
The path distribution corresponding to agent $B$ is obtained analogously. Substituting the resulting expression along with~(\ref{path_dist_A}) into~(\ref{prob_dist_paths_1_fact}), and thus obtained expression and~(\ref{prob_dist_paths_term_1}) into~(\ref{prob_dist_paths_1}) yields
\begin{eqnarray}
\nonumber & p \propto p({\bf r}_t | {\bf x}^A_t, {\bf x}^B_t) \; \delta({\bf x}^A_t-{\bf x}^A_{t^-}) \; \delta({\bf x}^B_t-{\bf x}^B_{t^-}) \times \\
& p({\bf x}^A_{1:t^-} | \mathcal{D}^A_{1:t^-}) \; p({\bf x}^B_{1:t^-} | \mathcal{D}^B_{1:t^-}) \label{prob_dist_paths_final}
\end{eqnarray}

A sequential inference of agents' paths is performed with particle filter (PF), directly derived from~(\ref{prob_dist_paths_final}). Namely, the agents' path particles up to time $t$ are sampled from proposal distribution defined as
\begin{equation}
q = \delta({\bf x}^A_t-{\bf x}^A_{t^-}) \; \delta({\bf x}^B_t-{\bf x}^B_{t^-}) \; p({\bf x}^A_{1:t^-} | \mathcal{D}^A_{1:t^-}) \; p({\bf x}^B_{1:t^-} | \mathcal{D}^B_{1:t^-})
\end{equation}
Thus, pose particles $\hat{\bf x}^A_{t,k}$ and $\hat{\bf x}^B_{t,k}$ of the two agents at time $t$ are obtained from the respective path posteriors $p({\bf x}^A_{1:t^-} | \mathcal{D}^A_{1:t^-})$ and $p({\bf x}^B_{1:t^-} | \mathcal{D}^A_{1:t^-})$. Each path posterior is represented with its own set of pose particles, evaluated using single-agent positioning algorithm outlined in Section~\ref{subsec:singe_agent_pose} such that the agents' particles representing their poses at time $t^-$ are $\{\hat{\bf x}^A_{t^-,k}\}_{k=1}^K$ and $\{\hat{\bf x}^B_{t^-,k}\}_{k=1}^K$, where we assume without loss of generality that both agents use the same number of particles $K$. Therefore, $\hat{\bf x}^A_{t,k}$ and $\hat{\bf x}^B_{t,k}$ are independently sampled from $\{\hat{\bf x}^A_{t^-,k}\}_{k=1}^K$ and $\{\hat{\bf x}^B_{t^-,k}\}_{k=1}^K$ with sampling distributions defined by the associated weights $\{w^A_{t^-,k}\}_{k=1}^K$ and $\{w^B_{t^-,k}\}_{k=1}^K$, respectively. In the case single-agent positioning algorithm resamples particles before each update step, the sampling distributions are uniform. The weight associated to the sampled pair of agents' pose particles, $\hat{\bf x}^A_{t,k}$ and $\hat{\bf x}^B_{t,k}$, is given as the ratio between the target $p$ and proposal distribution $q$ such that~\cite{Arulampalam_PF_02,Djuric_PF_03}
\begin{equation}
w_k = \frac{p}{q} \propto p({\bf r}_t | \hat{\bf x}^A_{t,k}, \hat{\bf x}^B_{t,k})
\end{equation}
Once $K$ pose particle pairs are sampled (so as to keep the number of particles unchanged), their weights are normalized so that $\sum_{k=1}^K w_k = 1$. Upon resampling pose particle pairs according to $\{w_k\}_{k=1}^K$, the resulting particles fuse all available measurements and represent agents' pose distributions upon their encounter.

The described approach for updating agents' pose particles at the encounter naturally admits a decentralized implementation across agents. Specifically, each agent performs a PF-based single-agent positioning using its odometry and ranging measurements until it comes relatively close to some other agent. At an encounter, the agents detect presence of each other, measure their relative pose and exchange their most recent pose particle sets. Each agent then updates its pose particle set on its own, using its particle set prior to the encounter, received particle set of the other agent and the relative pose measurement. It then continues to move in the area and perform single-agent positioning, initialized with the updated particle set. The pseudo-code of the proposed algorithm implemented on agent $A$ is given in Algorithm~\ref{alg}. The implementation for the other agent is analogous. 
\begin{algorithm}
	\caption{Proposed algorithm implementation on Agent $A$}
	\label{alg}
	\begin{algorithmic}[1]
		\State Inputs: $\mathcal{X}^A = \{\hat{\bf x}^A_{t^-,k}\}_{k=1}^K$, $\mathcal{X}^B = \{\hat{\bf x}^B_{t^-,k}\}_{k=1}^K$
		\State Inputs: Measurement of the agents' relative pose ${\bf r}_t$
		\State Initialize $\tilde{\mathcal{X}}^A = \mathcal{A}=\emptyset$
		\For{$k=1:K$}
			\State Uniformly sample $\tilde{\bf x}^A_{t,k} \sim \mathcal{X}^A$
			\State Uniformly sample $\tilde{\bf x}^B_{t,k} \sim \mathcal{X}^B$
			\State Compute $w_k \propto p({\bf r}_t \; | \; \tilde{\bf x}_{t,k}^A , \tilde{\bf x}_{t,k}^B)$
			\State Update $\mathcal{A} = \mathcal{A} \cup \{(\tilde{\bf x}^A_{t,k} , w_k)\} $
		\EndFor
		\State Normalize $w_k \leftarrow \frac{w_k}{\sum_{k=1}^K w_k}$
	  	\For{$l=1:K$}
			\State Draw $i$ with probability $w_i$
			\State Update $\tilde{\mathcal{X}}^A  = \tilde{\mathcal{X}}^A  \cup \tilde{\bf x}^A_{t,i}$
		\EndFor
		\State \Return $\tilde{\mathcal{X}}^A$	
	\end{algorithmic}
\end{algorithm}

In addition to admitting decentralized implementation, the proposed algorithm has some other favorable features. 
First, it is derived within the Bayesian inference framework and thus the particle updating scheme is optimal in the Bayesian sense.
Second, the computational complexity of the update step comprises of evaluating weights for pose particle pairs and is no larger than an update step in a conventional single-agent positioning. Third, the agents exchange their pose particles at the encounter, occurring when they are in the vicinity of each other, implying relatively mild requirements for communication bandwidth and coverage. Fourth, the algorithm seamlessly supports heterogeneous agents employing sensors of different capabilities. Specifically, as the above mathematical derivation suggests, an agent does not need any information about other agent's platform and/or sensors because this information is already implicitly contained in pose particles it receives from that agent. Fifth, the algorithm is scalable to more that two agents over their multiple encounters. Namely, an agent performs pose particle update using Algorithm~\ref{alg} at each encounter with any other agent. 
Sixth, the derived algorithm holds regardless of the type of measurements used for single-agent positioning, as long as the poses are represented with particle clouds. Finally, the performance benefits of the proposed algorithm are experimentally validated, which is described in the following part. 
\begin{figure*}
\begin{minipage}{\textwidth}
\centering
\begin{subfigure}[b]{0.3\textwidth}
\centering
\includegraphics[scale=0.37]{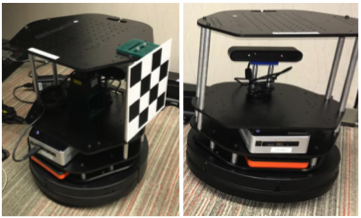}
\caption{Robots (left: $A$, right: $B$)} 
\label{robots_pic}
\end{subfigure}
\begin{subfigure}[b]{0.3\textwidth}
\centering
\includegraphics[scale=0.07]{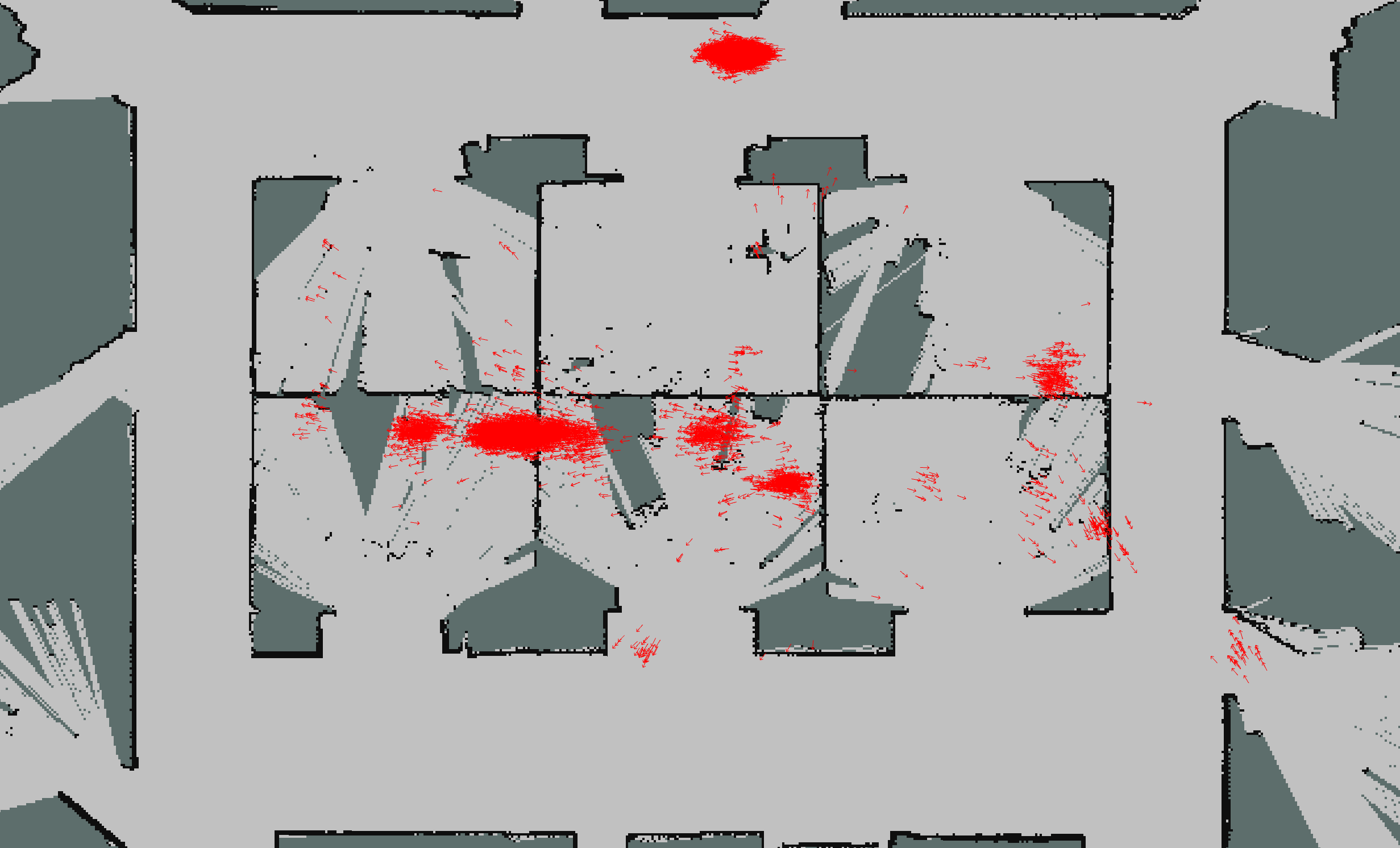}
\caption{Robot $A$ prior to the encounter} 
\label{robot_A_before}
\end{subfigure}
\begin{subfigure}[b]{0.3\textwidth}  
\centering 
\includegraphics[scale=0.07]{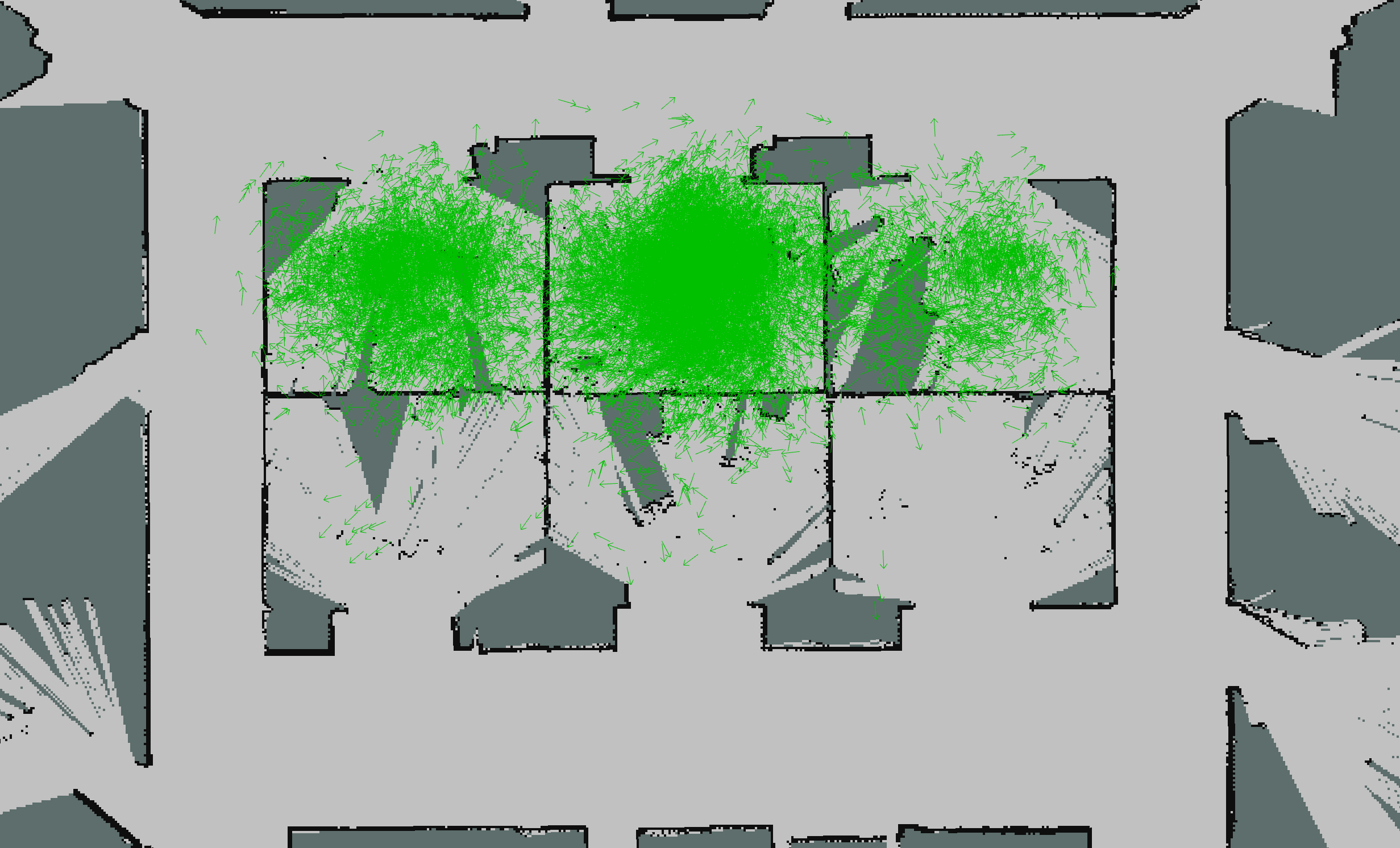}
\caption{Robot $B$ prior to the encounter}
\label{robot_B_before}
\end{subfigure}
\vskip\baselineskip
\begin{subfigure}[b]{0.3\textwidth}
\centering
\includegraphics[scale=0.18]{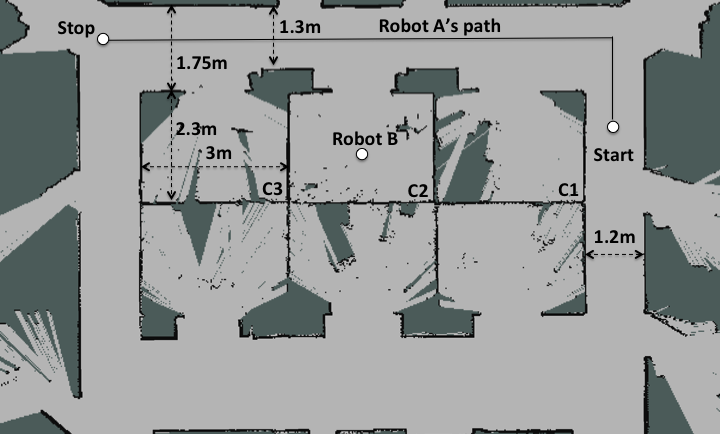}
\caption{Experimental setup} 
\label{area_map}
\end{subfigure}
\begin{subfigure}[b]{0.3\textwidth}
\centering
\includegraphics[scale=0.07]{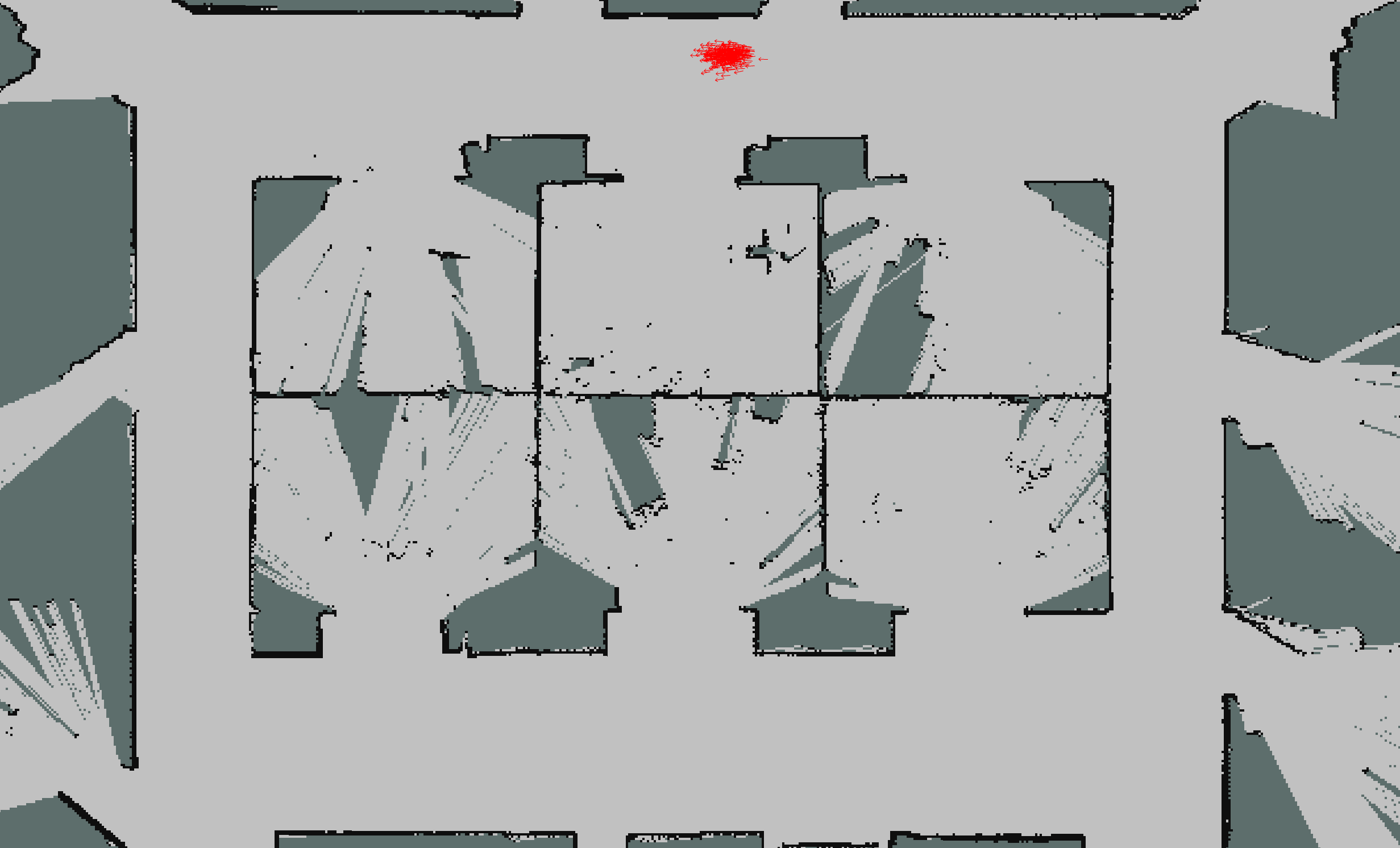}
\caption{Robot $A$ upon the encounter} 
\label{robot_A_after}
\end{subfigure}
\begin{subfigure}[b]{0.3\textwidth}  
\centering 
\includegraphics[scale=0.07]{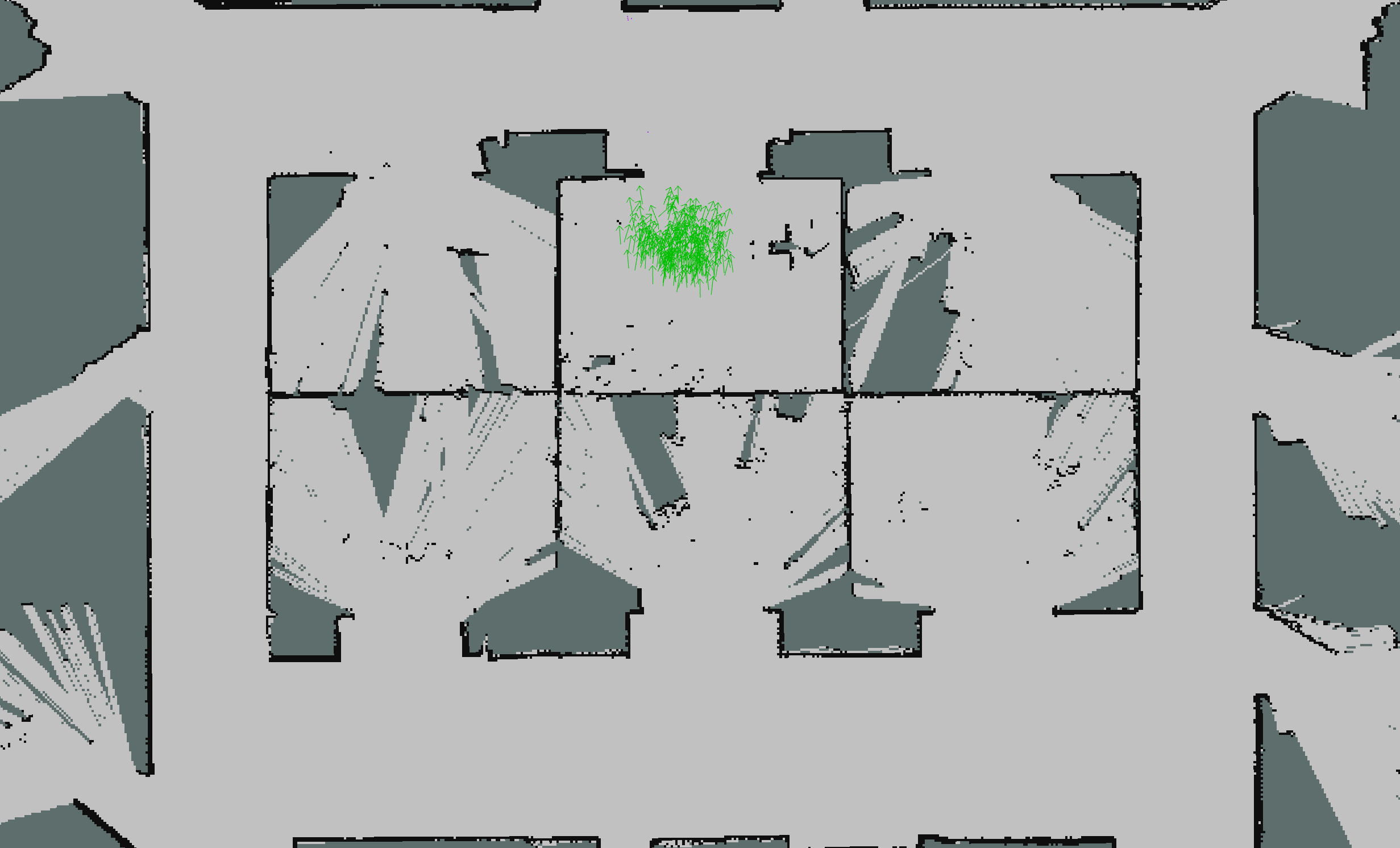}
\caption{Robot $B$ upon the encounter}
\label{robot_B_after}
\end{subfigure}
\caption{Experimental robots, setup and pose particles of robots $A$ and $B$ prior and upon their encounter in one experiment.}
\label{particle_sets}
\end{minipage}
\end{figure*}

\section{Experimental Validation}

We use two open source Turtlebot platforms{\footnote{https://www.turtlebot.com}} shown in Fig.~\ref{robots_pic} for the experimental validation.
These robots, equipped with full odometry and RGBD sensors, provide odometry ${\bf u}_t$ and ranging measurements ${\bf z}_t$, where the measurement models are from~\cite[Ch 5 and 6]{Thrun_book_05}. 
To enable cooperation, robot $A$ has a rigidly attached $3\times 3$ checkerboard pattern to its body so that robot $B$ can detect it based on the optical images from the RGBD sensor. 
Upon detecting presence of robot $A$, robot $B$ performs relative pose measurement in two steps. First, the location of the center of the checkerboard pattern is obtained from the point cloud created from the depth measurements of the RGBD sensor. Then, the relative pose is estimated by extracting the point cloud of the checkerboard corners and estimating the surface normal using OpenPCL~\cite{openPCL}. Robot $B$ then sends the relative pose measurement to robot $A$ and the two robots exchange their pose particles. 
As justified in~\cite{Fox_MultiRobotLocalization_00} and used in~\cite{Prorok_IROS_11,Prorok_ICRA_12}, we assume the measurement errors of the relative distance $r$ and heading $\theta$ are independent and zero-mean Gaussian distributed with standard deviations $\sigma_r=0.1$\;m and $\sigma_{\phi}=10^o$, selected based on test results of the described relative pose estimation pipeline.

An occupancy grid map of the experimental area with $2.5$\;cm cell resolution is available to both robots. As shown in Fig.~\ref{area_map}, robot $B$ is stationary and robot $A$ moves along a corridor such that they meet when robot $A$ passes by the cubicle where robot $B$ is located. The pose particles of robot $B$ are prior to the encounter with robot $A$ shown in Fig.~\ref{robot_B_before}. Since robot $A$ has no prior knowledge of its initial pose, its particles have not yet converged by the time the robots meet, as shown in Fig.~\ref{robot_A_before}. However, even in such a case where robots have little knowledge of their poses prior to the encounter, the information fusion happening at the encounter according to the proposed algorithm results in updated particles that are fairly confined around the true poses of the corresponding robots, as shown in Figs~\ref{robot_A_after} and~\ref{robot_B_after}.

\begin{figure}[thpb]
\centering
\includegraphics[scale=0.25]{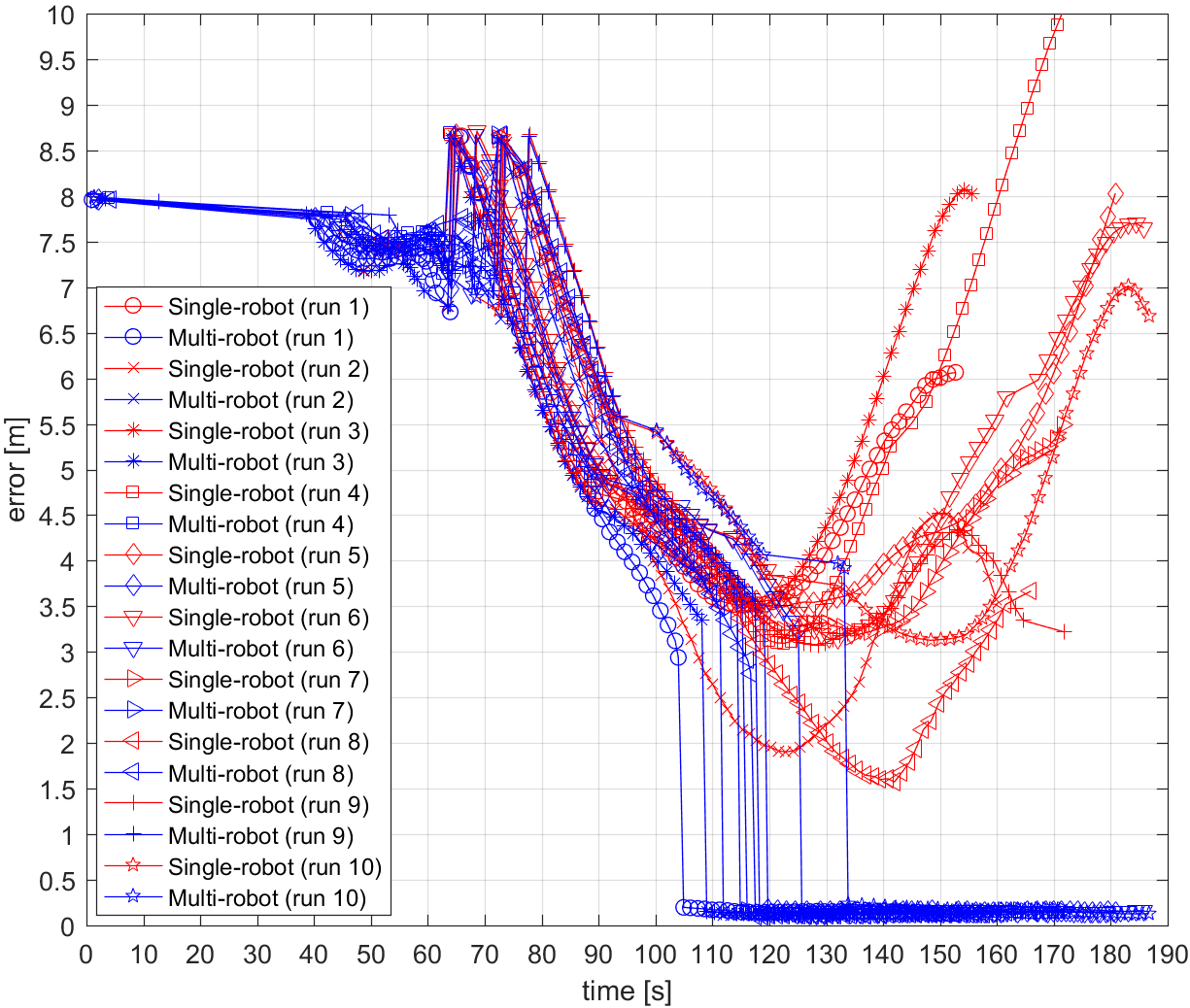}
\caption{Positioning errors across ten experimental runs.}
\label{err_plots}
\end{figure}

The proposed algorithm is benchmarked against a single-agent positioning algorithm~\cite{AMCL_alg}, conventionally used in robotics. Overall, the two algorithms perform global positioning of robot $A$, and are run synchronously over the same measurements. The ground truth positions of robot $A$ are estimated from averaging pose particles resulting from running yet another instance of the conventional single-agent positioning algorithm, initialized with the correct starting position of robot $A$ and whose ranging measurements are provided by a high-resolution lidar. The comparison of how positioning error of the two algorithms evolve over time across ten independent experimental runs is shown in Fig.~\ref{err_plots}. 
As shown, the error corresponding to the proposed algorithm drops and stays below $0.2$\;m (and in most cases below $0.1\;$m) as soon as the robots meet in all ten experimental runs. Not surprisingly, the encounters in different runs happen at different time instants. The shown results validate that the proposed algorithm yields better error performance and faster convergence of particles, compared to the single-agent positioning case. 
In addition, due to a symmetry in the experimental area around the vertical axis passing through the middle of the area, the error corresponding to the single-agent positioning algorithm increases, indicating that the particles diverge after robot $A$ passes by the cubicle where robot $B$ is located. In contrast, the proposed algorithm is immune to this symmetry and 
is able to accurately track robot $B$ as it moves down the second half of its route.

\section{Conclusion}
We describe an algorithm for cooperative positioning of multiple agents. The proposed algorithm has a relatively mild communication bandwidth and coverage requirements, and is implemented as a fully decentralized particle filter with the same complexity as a single-agent positioning method. The experiments confirm better accuracy and faster convergence compared to the case when agents do not cooperate, and show its immunity to a symmetric area where a single-agent positioning fails.

\newpage


%





\ifCLASSOPTIONcaptionsoff
  \newpage
\fi


\begin{thebibliography}{1}

\bibitem{CITS_1} L. Chen and C. Englund, ``Cooperative ITS - EU Standards to Accelerate Cooperative Mobility", {\it International Conference on Connected Vehicles and Expo (IC-CVE)}, Nov. 2014, pp. 681-686.
\bibitem{CITS_2} K. Sjoberg, P. Andres, T. Buburuzan and A. Brakemeier, ``Cooperative Intelligent Transport Systems in Europe: Current Deployment Status and Outlook," {\it IEEE Vehicular Technology Magazine}, vol. 12, no. 2, pp. 89-97, Jun. 2017.

\bibitem{urban_canyon} E. F. N. A. Boukerche, H. A. B. F. Oliveira and A. A. Loureiro, ``Vehicular Ad Hoc Networks: A New Challenge for Localization-Based Systems," {\it Computer Communications,} vol. 31, no. 12, pp. 2838-2849, Jul. 2008.

\bibitem{lit_CP_GPS_1} M. Rohani, D. Gingras, V. Vigneron and D. Gruyer, ``A New Decentralized Bayesian Approach for Cooperative Vehicle Localization Based on Fusion of GPS and VANET Based Inter-Vehicle Distance Measurement", {\it IEEE Intelligent Transportation Systems Magazine,} vol. 7, no. 2, pp. 85-95, Apr. 2015.
\bibitem{lit_CP_GPS_2} R. Parker and S. Valaee, ``Vehicular Node Localization Using Received-Signal-Strength Indicator", {\it IEEE Transactions on Vehicular Technology,} vol. 56, no. 6, pp. 3371-3380, Nov. 2007.
\bibitem{lit_CP_GPS_3} G. Soatti, M. Nicoli, S. Savazzi and U. Spagnolini, ``Consensus-Based Algorithms for Distributed Network-State Estimation and Localization", {\it IEEE Transactions on Signal and Information Processing over Networks,} vol. 3, no. 2, pp. 430-444, Jun. 2017.
\bibitem{lit_CP_GPS_4} H. Wymeersch, J. Lien and M. Z. Win, ``Cooperative Localization in Wireless Networks", {\it Proceedings of the IEEE,} vol. 97, no. 2, pp. 427-450, Feb. 2009.
\bibitem{lit_CP_GPS_5} N. Alam, A. T. Balaei and A. G. Dempster, ``A DSRC Doppler-Based Cooperative Positioning Enhancement for Vehicular Networks With GPS Availability", {\it IEEE Transactions on Vehicular Technology,} vol. 60, no. 9, pp. 4462-4470, Nov. 2011.

\bibitem{BayesianCP_PDA_separate} G. Soatti, M. Nicoli, N. Garcia, B. Denis, R. Raulefs and H. Wymeersch,``Implicit Cooperative Positioning in Vehicular Networks", {\it IEEE Transactions on Intelligent Transportation Systems,} 2018.
\bibitem{BayesianCP_PDA_joint} M. Brambilla, G. Soatti and M. Nicoli, ``Precise Vehicle Positioning by Cooperative Feature Association and Tracking in Vehicular Networks," {\it 2018 IEEE Statistical Signal Processing Workshop (SSP)}, Freiburg, 2018, pp. 648-652.

\bibitem{Saeedi_review_16} S. Saeedi, M. Trentini, M. Seto and H. Li, ``Multiple-robot Simultaneous Localization and Mapping: A review", {\it Journal of Field Robotics}, vol. 33, no. 1, pp. 3--46, Jan. 2016.

\bibitem{AMCL_alg} D. Fox, ``Adapting the Sample Size in Particle Filters through KLD-sampling", {\it International Journal of Robotics Research}, vol. 22, no. 12, pp. 985--1003, Dec. 2003.
\bibitem{SLAM} G. Grisetti, C. Stachniss and W. Burgard, ``Improved Techniques for Grid Mapping with Rao-Blackwellized Particle Filters", {\it IEEE Transactions on Robotics}, vol. 23, no. 1, pp. 34--46, Feb. 2007.

\bibitem{Djuric_PF_03} P. M. Djuric, J. H. Kotecha, J. Zhang, Y. Huang, T. Ghirmai, M. F. Bugallo and J. Miguez, ``Particle Filtering", {\it IEEE Signal Processing Magazine}, vol. 20, no. 5, pp. 19--38, Sept. 2003.
\bibitem{Arulampalam_PF_02} M. S. Arulampalam, S. Maskell, N. Gordon and T. Clapp, ``A tutorial on particle filters for online nonlinear/non-Gaussian Bayesian tracking", {\it IEEE Transactions on Signal Processing}, vol. 50, no. 2, pp. 174--188, Feb. 2002.

\bibitem{red_odom_err_1} J. Borenstein, ``Control and kinematic design of multi-degree-of-freedom robots with compliant linkage", {\it IEEE Transactions on Robotics and Automation}, 1995.
\bibitem{red_odom_err_2} R. Kurazume and N. Shigemi, ``Cooperative positioning with multiple robots", In {\it Proc. of the IEEE/RSJ International Conference on Intelligent Robots and Systems (IROS)}, 1994.
\bibitem{red_odom_err_3} I. M. Rekleitis, G. Dudek, and E. Milios, ``Multi-robot exploration of an unknown environment, efficiently reducing the odometry error", In {\it Proc. of the International Joint Conference on Artificial Intelligence (IJCAI)}, 1997.

\bibitem{Kalman_1} S. I. Roumeliotis and G. A. Bekey, ``Distributed multirobot localization", In {\it Transactions on Robotics and Automation}, vol. 18, no. 5, pp. 781--795, 2002.
\bibitem{Kalman_2} E. D. Nerurkar, S. I. Roumeliotis and A. Martinelli, ``Distributed maximum a posteriori estimation for multi-robot cooperative localization", In {\it Proc. of the IEEE International Conference on Robotics and Automation (ICRA)}, pp. 1402--1409, May 2009.
\bibitem{Kalman_3} L. C. Carrillo-Arce, E. D. Nerurkar, J. L. Gordillo and S. I. Roumeliotis, ``Decentralized multi-robot cooperative localization using covariance intersection", In {\it IEEE/RSJ International Conference on Intelligent Robots and Systems (IROS)}, pp. 1412--1417, 2013.

\bibitem{opt_based_1} S. Safavi and U. Khan, ``An Opportunistic Linear Convex Algorithm for Localization in Mobile Robot Networks", {\it  IEEE Transactions on Robotics}, vol. 33, no. 4, Aug. 2017.
\bibitem{opt_based_2} A. Franchi, G. Oriolo and P. Stegagno, ``Mutual localization in multi-robot systems using anonymous relative measurements", {\it The International Journal of Robotics Research}, vol. 32, no. 11, pp. 1302--1322, 2013.

\bibitem{Carlone_SLAM_10} L. Carlone, M. K. Ng, J. Du, B. Bona, and M. Indri, ``Rao-Blackwellized Particle Filters Multi Robot SLAM with Unknown Initial Correspondences and Limited Communication", In {\it Proc. of the IEEE International Conference on Robotics and Automation (ICRA)}, pp. 243--249, May 2010.
\bibitem{Howard_MultiRobotSLAM_06} A. Howard, ``Multi-robot Simultaneous Localization and Mapping using Particle Filters", {\it The International Journal of Robotics Research}, vol. 25, no. 12, pp. 1243-1256, Dec. 2006.

\bibitem{CL_tracking} A. Ahmad, G. Lawless and P. Lima, ``An Online Scalable Approach to Unified Multirobot Cooperative Localization and Object Tracking", {\it IEEE Transactions on Robotics}, vol. 33, no. 5, pp. 1184--1199, Oct. 2017.

\bibitem{Fox_MultiRobotLocalization_00} D. Fox, W. Burgard, H. Kruppa and S. Thrun, ``A Probabilistic Approach to Collaborative Multi-Robot Localization", {\it Autonomous Robots}, vol. 8, no. 3, pp. 325--344, Jun 2000.
\bibitem{Prorok_IROS_11} A. Prorok and A. Martinoli, ``A reciprocal sampling algorithm for lightweight distributed multi-robot localization", In {\it Proc. of the IEEE/RSJ International Conference on Intelligent Robots and Systems (IROS)}, pp. 3241--3247, 2011.
\bibitem{Prorok_ICRA_12} A. Prorok, A. Bahr and A. Martinoli, ``Low-cost collaborative localization for large-scale multi-robot systems", In {\it IEEE International Conference on Robotics and Automation (ICRA)}, pp. 4236-4241, 2012.

\bibitem{Thrun_book_05} S. Thrun, W. Burgard, and D. Fox, ``Probabilistic Robotics (Intelligent Robotics and Autonomous Agents)", {\it The MIT Press}, 2005.

\bibitem{openPCL} R. B. Rusu and S. Cousins, ``3D is here: Point Cloud Library (PCL)", In {\it Proc. of the IEEE International Conference on Robotics and Automation (ICRA)}, May 2011.

\bibitem{Ristic_book_04} B. Ristic, S. Arulampalm and N. Gordon, ``Beyond the Kalman filter : particle filters for tracking applications", {\it Artech House}, 2004.

\end{thebibliography}
\end{document}